\documentclass[10pt,letterpaper]{article}
\usepackage[top=0.775in,left=2.75in,footskip=0.75in]{geometry}

\usepackage{amsmath,amssymb}

\usepackage{changepage}

\usepackage[utf8x]{inputenc}

\usepackage{textcomp,marvosym}

\usepackage{cite}

\usepackage{nameref,hyperref}
\usepackage{url}
\usepackage{breakurl}

\usepackage{microtype}
\DisableLigatures[f]{encoding = *, family = * }

\usepackage[table]{xcolor}

\usepackage{array}

\newcolumntype{+}{!{\vrule width 2pt}}

\newlength\savedwidth

\newcommand\thickhline{\noalign{\global\savedwidth\arrayrulewidth\global\arrayrulewidth 2pt}%
\hline
\noalign{\global\arrayrulewidth\savedwidth}}

\raggedright
\usepackage[aboveskip=1pt,labelfont=bf,labelsep=period,justification=raggedright,singlelinecheck=off]{caption}

\makeatletter
\renewcommand{\@biblabel}[1]{\quad#1.}
\makeatother

\usepackage{lastpage,fancyhdr,graphicx}
\usepackage{epstopdf}
\usepackage{tikz}
\fancyheadoffset[L]{2.25in}
\fancyfootoffset[L]{2.25in}
\newcommand{\OF}{$\infty$ }
\newcommand{\Ref}[1]{\textit{\nameref{#1}}}

\begin{document}
\vspace*{0.2in}

\begin{flushleft}
{\Large
\textbf\newline{A hierarchical loss and its problems when classifying non-hierarchically\\\phantom{---}} 
}
Cinna Wu\textsuperscript{1},
Mark Tygert\textsuperscript{1*},
Yann LeCun\textsuperscript{2}
\\
\bigskip
\textbf{1} Facebook, 1 Facebook Way, Menlo Park, CA 94025
\\
\textbf{2} Facebook, 770 Broadway, New York, NY 10003
\\
\bigskip

%
%





* corresponding author: tygert@fb.com

\end{flushleft}
\section*{Abstract}
Failing to distinguish between a sheepdog and a skyscraper should be worse
and penalized more than failing to distinguish between a sheepdog and a poodle;
after all, sheepdogs and poodles are both breeds of dogs.
However, existing metrics of failure (so-called ``loss'' or ``win'')
used in textual or visual classification/recognition via neural networks
seldom leverage a-priori information, such as a sheepdog being more similar
to a poodle than to a skyscraper.
We define a metric that, inter alia, can penalize failure to distinguish
between a sheepdog and a skyscraper more than failure to distinguish
between a sheepdog and a poodle. Unlike previously employed possibilities,
this metric is based on an ultrametric tree associated
with any given tree organization into a semantically meaningful hierarchy
of a classifier's classes. An ultrametric tree is a tree
with a so-called ultrametric distance metric
such that all leaves are at the same distance from the root.
Unfortunately, extensive numerical experiments indicate that
the standard practice of training neural networks
via stochastic gradient descent with random starting points
often drives down the hierarchical loss nearly as much
when minimizing the standard cross-entropy loss
as when trying to minimize the hierarchical loss directly.
Thus, this hierarchical loss is unreliable as an objective
for plain, randomly started stochastic gradient descent to minimize;
the main value of the hierarchical loss may be merely as a meaningful metric
of success of a classifier.



\section*{Introduction}
\label{intro}

Metrics for classifier accuracy used in the neural network methods
of~\cite{lecun-bengio-hinton} seldom account
for semantically meaningful organizations of the classes;
these metrics neglect, for instance, that sheepdogs and poodles are dogs,
that dogs and cats are mammals, that mammals, birds, reptiles, amphibians,
and fish are vertebrates, and so on.
Below, we define a metric --- the amount of the ``win'' or ``winnings''
for a classification --- that accounts for a given organization
of the classes into a tree. During an optimization
(also known as ``training''), we want to maximize the win or, equivalently,
minimize the ``loss'' (loss is the negative of the win).
We contrast the hierarchical win to the existing universal standard,
the cross-entropy loss discussed by~\cite{lecun-bengio-hinton}
(and by many others).
We caution that several of our experiments indicate
that plain stochastic gradient descent optimization
with random starting points can get stuck in local optima; even so,
the hierarchical win can serve as a good metric of success or figure of merit
for the accuracy of the classifier.

The approach detailed below is a special case
of the general methods of~\cite{cai-hofmann},
\cite{kosmopoulos-partalas-gaussier-paliouras-androutsopoulos},
and their references. The particular special cases discussed by them
and by~\cite{binder-kawanabe-brefeld, chang-lee, costa-lorena-carvalho-freitas,
deng-berg-li-li1, deng-berg-li-li2, wang-zhou-liew}
allocate different weights to different leaves,
not leveraging ``ultrametric trees'' as in the present paper
(an ultrametric tree is a tree with a so-called ultrametric distance metric
such that all leaves are at the same distance from the root,
as in the phylogenetics discussed, for example, by~\cite{campbell}).
The distinguishing characteristic of the distance metric constructed below
is that all leaves are at the same distance from the root,
as in an ultrametric tree; this provides a definition and measure
of successful classification that is different from other options.

A related topic is hierarchical classification,
as reviewed by~\cite{silla-freitas} (with further use more recently
by~\cite{kosmopoulos-paliouras-androutsopoulos} and \cite{redmon-farhadi},
for example); however, the present paper considers
only classification into a given hierarchy's finest-level classes.
The design options for classification into only the finest-level classes
are more circumscribed, yet such classification is easier to use, implement,
and interface with existing codes.
The existing standard cross-entropy is the chief alternative in the setting
of the present paper (that is, for classification
into only the finest-level classes).
Cross-entropy loss is the negative of the logarithm of our hierarchical win
when the hierarchy is ``flat,'' that is,
when the hierarchy is the degenerate case in which all classes
are leaves attached to the same root.
Extensive experiments demonstrate the advantages of hierarchical loss
in comparison to the conventional cross-entropy.

Hierarchical classification, such as that reviewed by~\cite{silla-freitas},
is more sophisticated (and potentially more powerful) than what the present
article considers. The present paper only changes the metric of success
(and objective function), not altering the procedure for classification ---
hierarchical procedures for classification are often avoided
due to complications of systems development and systems programming.
We were tasked with changing the figure of merit for measuring success
of a classifier without being allowed to pursue more ambitious,
full-blown hierarchical processes for classification.
The actual real-world motivation for the hierarchical loss
of the present article was in recommending places to users,
where (for example) sending users to a Vietnamese restaurant
when they request a Thai restaurant should be penalized less
than sending them to a park instead --- different kinds of restaurants
are more similar to each other than to a park.

We warn that minimizing the hierarchical loss directly
via the usual stochastic gradient descent with random starting points
was often fruitless even in relatively simple numerical experiments
reported below.
The complexity of systems deployed in practice typically makes
hierarchical procedures for optimization very hard to implement, yet
altering only the objective of the optimization has frustratingly little effect
on the hierarchical loss during several tests in Section~\Ref{numex} below.

The remainder of the present paper has the following structure:
Section~\Ref{methods} constructs the hierarchical loss and win.
Section~\Ref{numex} illustrates and evaluates the hierarchical loss and win
via several numerical experiments.
Section~\Ref{conclusion} draws conclusions and proposes directions
for future work.

\section*{Methods}
\label{methods}

Concretely, suppose that we want to classify each input into one
of many classes, and that these classes are the leaves
in a tree which organizes them into a semantically meaningful hierarchy.
Suppose further that a classifier maps an input
to an output probability distribution over the leaves,
hopefully concentrated on the leaf corresponding to the correct class
for the input.
We define the probability of any node in the tree to be
the sum of the probabilities of all leaves falling under the node
(the node represents an aggregated class consisting of all these leaves);
a leaf falls under the node if the node is on the path
from the root to the leaf.
We then define the amount of the ``win'' or ``winnings'' to be
the weighted sum (with weights as detailed shortly) of the probabilities
of the nodes along the path from the root to the leaf corresponding
to the correct class.

To calculate the win, we sum across all nodes on the path
from the root to the leaf corresponding to the correct class,
including both the root and the leaf,
weighting the probability at the first node (that is, at the root) by $1/2$,
weighting at the second node by $1/2^2$,
weighting at the third node by $1/2^3$,
weighting at the fourth node by $1/2^4$,
and so on until the final leaf.
We then add to this sum the probability at the final leaf,
weighted with the same weight as in the sum,
that is, we double-count the final leaf.
We justify the double-counting shortly.

To compute the probability of each node
given the probability distribution over the leaves,
we propagate the leaf probabilities through the tree as follows.
We begin by storing a zero at each node in the tree.
Then, for each leaf, we add the probability associated with the leaf
to the value stored at each node on the path from the root to the leaf,
including at the root and at the leaf.
We save these accumulated values as the propagated probabilities
(storing the value 1 at the root ---
the sum of all the probabilities is 1 of course).

Thus, if the probability of the final leaf is 1,
then the win is 1. The win can be as large as 1
(this happens when the classification is completely certain and correct)
or as small as 1/2 (this happens when the classification
is as wrong as possible). The win being 1 whenever the probability
of the final leaf is 1 --- irrespective of which is the final leaf ---
means that the weights form an ``ultrametric tree,''
as in the phylogenetics discussed, for example, by~\cite{campbell}.
This justifies double-counting the final leaf.

Figs~\ref{fishhier}--\ref{fishlast} illustrate the evaluation
of the hierarchical win via examples.
The following two subsections summarize in pseudocode
the algorithms for propagating the probabilities
and for calculating the win, respectively
(the latter algorithm runs the former as its initial step).

\begin{figure}
\begin{center}
\begin{tikzpicture}[level distance=.5in,
  level 1/.style={sibling distance=1in},
  level 2/.style={sibling distance=1in},
  every text node part/.style={align=center}]
\node{acipenseriformes}
  child {node {paddlefish}}
  child {node {sturgeon}
    child {node {acipenser}
      child {node {acipenser\\(oxyrinchus)}}
      child {node {acipenser\\(other)}}
    }
    child {node {huso}}
    child {node {pseudo-\\scaphirhynchus}}
    child {node {scaphirhynchus}}
    child {node {sturgeon\\(other)}}
  };
\end{tikzpicture}
\end{center}
\caption{This is a hierarchy of types of fish (but not the whole hierarchy
from Subsection~\Ref{fishsec} of Section~\Ref{numex}).}
\label{fishhier}
\end{figure}

\begin{figure}
\begin{center}
\begin{tikzpicture}[level distance=.5in,
  level 1/.style={sibling distance=1in},
  level 2/.style={sibling distance=1in},
  every text node part/.style={align=center},
  emph/.style={edge from parent/.style={very thick,draw}},
  norm/.style={edge from parent/.style={thin,draw}}]
\node{acipenseriformes\\+1/2}
  child {node {paddlefish}}
  child[emph] {node {sturgeon\\+1/4}
    child[emph] {node {acipenser}
      child[emph] {node {acipenser\\(oxyrinchus)}}
      child[norm] {node {acipenser\\(other)}}
    }
    child[norm] {node[draw] {huso}}
    child[norm] {node {pseudo-\\scaphirhynchus}}
    child[norm] {node {scaphirhynchus}}
    child[norm] {node {sturgeon\\(other)}}
  };
\end{tikzpicture}
\end{center}
\caption{This depicts the hierarchy of Fig~\ref{fishhier} when the correct
target leaf is ``acipenser (oxyrinchus)'' but the leaf predicted
with probability 1 (the ``single best class'' of Subsection~\Ref{inbest}) is
``huso'' (as indicated with a box around ``huso'').
In this case, the hierarchical win is 1/2 + 1/4 = 3/4, as ``acipenseriformes''
and ``sturgeon'' are the nodes on both the path
from the root ``acipenseriformes'' to ``acipenser (oxyrinchus)'' and the path
from the root ``acipenseriformes'' to ``huso.''}
\label{fish2}

\vspace{1in}

\end{figure}

\begin{figure}
\begin{center}
\begin{tikzpicture}[level distance=.5in,
  level 1/.style={sibling distance=1in},
  level 2/.style={sibling distance=1in},
  every text node part/.style={align=center},
  emph/.style={edge from parent/.style={very thick,draw}},
  norm/.style={edge from parent/.style={thin,draw}}]
\node{acipenseriformes\\+1/2}
  child {node {paddlefish}}
  child[emph] {node {sturgeon\\+1/4}
    child[emph] {node {acipenser\\+1/8}
      child[emph] {node {acipenser\\(oxyrinchus)}}
      child[norm] {node[draw] {acipenser\\(other)}}
    }
    child[norm] {node {huso}}
    child[norm] {node {pseudo-\\scaphirhynchus}}
    child[norm] {node {scaphirhynchus}}
    child[norm] {node {sturgeon\\(other)}}
  };
\end{tikzpicture}
\end{center}
\caption{This depicts the hierarchy of Fig~\ref{fishhier} when the correct
target leaf is ``acipenser (oxyrinchus)'' but the leaf predicted
with probability 1 (the ``single best class'' of Subsection~\Ref{inbest}) is
``acipenser (other)'' (as indicated with a box around ``acipenser (other)'').
In this case, the hierarchical win is 1/2 + 1/4 + 1/8 = 7/8,
as ``acipenseriformes,'' ``sturgeon,'' and ``acipenser'' are the nodes on both
the path from the root ``acipenseriformes'' to ``acipenser (oxyrinchus)'' and
the path from the root ``acipenseriformes'' to ``acipenser (other).''}
\label{fish3}
\end{figure}

\begin{figure}
\begin{center}
\begin{tikzpicture}[level distance=.5in,
  level 1/.style={sibling distance=1in},
  level 2/.style={sibling distance=1in},
  every text node part/.style={align=center},
  emph/.style={edge from parent/.style={very thick,draw}},
  norm/.style={edge from parent/.style={thin,draw}}]
\node{acipenseriformes\\+1/2}
  child {node {paddlefish}}
  child[emph] {node {sturgeon\\+1/4}
    child[emph] {node {acipenser\\+1/8}
      child[emph] {node[draw] {acipenser\\(oxyrinchus)\\+1/8}}
      child[norm] {node {acipenser\\(other)}}
    }
    child[norm] {node {huso}}
    child[norm] {node {pseudo-\\scaphirhynchus}}
    child[norm] {node {scaphirhynchus}}
    child[norm] {node {sturgeon\\(other)}}
  };
\end{tikzpicture}
\end{center}
\vspace{-.1in}
\caption{This depicts the hierarchy of Fig~\ref{fishhier} when the correct
target leaf is ``acipenser (oxyrinchus),'' the same as the leaf predicted
with probability 1 (the ``single best class'' of Subsection~\Ref{inbest}).
In this case, the hierarchical win is 1/2 + 1/4 + 1/8 + 1/8 = 1
(as with all correct predictions);
the final node is double-counted at 1/8 = 1/16 + 1/16.}
\label{fishlast}
\end{figure}

\subsection*{An algorithm for propagating probabilities}
\label{probprop}

{
\small
\begin{flushleft}
{

{\bf Input}: a discrete probability distribution over the leaves of a tree

\smallskip

{\bf Output}: a scalar value (the total probability) at each node in the tree

\smallskip

{\bf Procedure}:

{\bf Store} the value 0 at each node in the tree.

{\bf For} each leaf,

\quad {\bf for} each node in the path from the root to the leaf
(including both the root and the leaf),

\quad \quad {\bf add} to the value stored at the node the probability
of the leaf.

}
\end{flushleft}
}

\subsection*{An algorithm for computing the ``win'' or ``winnings''}
\label{win}

{
\small
\begin{flushleft}
{

{\bf Input}: two inputs, namely (1) a discrete probability distribution
over the leaves of a tree, and (2) which leaf corresponds
to a completely correct classification

\smallskip

{\bf Output}: a single scalar value (the ``win'' or ``winnings'')

\smallskip

{\bf Procedure}:

{\bf Run} \Ref{probprop} to obtain a scalar at each node in the tree.

{\bf Store} the value 0 in an accumulator.

{\bf Define} $\ell$ to be the number of nodes
in the path from the root (node 1) to the leaf (node $\ell$).

{\bf For} $j = 1$,~$2$, \dots, $\ell$,

\quad {\bf add} to the accumulated value $2^{-j}$ times the value stored
at the path's $j$th node.

{\bf Add} to the accumulated value $2^{-\ell}$ times the value stored
at the path's $\ell$th node (the leaf).

{\bf Return} the final accumulated value.

}
\end{flushleft}
}

\subsection*{Calculation of gradients}

To facilitate optimization via gradient-based methods,
we now detail how to compute the gradient of the hierarchical win
with respect to the input distribution:
Relaxing the constraint that the distribution over the leaves
be a probability distribution, that is, that the ``probabilities''
of the leaves be nonnegative and sum to 1, the algorithms specified
in the preceding two subsections yield
a value for the win as a function of any distribution over the leaves. 
Without the constraint that the distribution over the leaves
be a probability distribution, the win is actually a linear function
of the distribution, that is, the win is the dot product between
the input distribution and another vector (where this other vector depends
on which leaf corresponds to the correct classification);
the gradient of the win with respect to the input is therefore just the vector
in the dot product.
An entry of this gradient vector, say the $j$th entry,
is equal to the win for the distribution over the leaves that consists
of all zeros except for one value of one on the $j$th leaf;
this win is equal to the sum of the series
$1/2 + 1/2^2 + 1/2^3 + 1/2^4 + \dots$, truncated to the number of terms
equal to the number of nodes for which the path from the root
to the correct leaf and the path from the root to the $j$th leaf coincide
(or not truncated at all if the $j$th leaf happens to be the same as the leaf
for a correct classification).

\subsection*{Choosing a single best class}
\label{inbest}
If forced to choose a single class corresponding to a leaf
of the given hierarchy (rather than classifying into a probability distribution
over all leaves) for the final output of the classification,
we first identify the node having greatest probability among all nodes
(including leaves) at the coarsest level, then the node
having greatest probability among all nodes (including leaves) falling
under the first node selected, then the node having greatest probability
among all nodes (including leaves) falling under the second node selected,
and so on, until we select a leaf. The leaf we select corresponds
to the class we choose.
The following subsection summarizes in pseudocode this procedure for selecting
a single best leaf.

\subsection*{An algorithm for choosing the single best leaf}
\label{top-down}

{
\small
\begin{flushleft}
{

{\bf Input}: a discrete probability distribution over the leaves of a tree

\smallskip

{\bf Output}: a single leaf of the tree

\smallskip

{\bf Procedure}:

{\bf Run} \Ref{probprop} to obtain a scalar at each node in the tree.

{\bf Move} to the root.

{\bf Repeating until} at a leaf,

\quad {\bf follow} (to the next finer level)
the branch containing the greatest among all scalar values

\quad \quad stored at this next finer level in the current subtree.

{\bf Return} the final leaf.

}
\end{flushleft}
}

\subsection*{Logarithms when using a softmax or independent samples}
\label{softmax}

In order to provide appropriately normalized results,
the input to the hierarchical loss needs to be a (discrete)
probability distribution, not just an arbitrary collection of numbers.
A ``softmax'' provides a good, standard means of converting any collection
of real numbers into a proper probability distribution.
Recall that the softmax of a sequence $x_1$,~$x_2$, \dots, $x_n$
of $n$ numbers is the normalized sequence $\exp(x_1)/Z$, $\exp(x_2)/Z$, \dots,
$\exp(x_n)/Z$, where $Z = \sum_{k=1}^n \exp(x_k)$.
Notice that each of the normalized numbers lies between 0 and 1,
and the sum of the numbers in the normalized sequence is 1 ---
the normalized sequence is a proper probability distribution.

When generating the probability distribution over the leaves via a softmax,
we should optimize based on the logarithm of the ``win'' introduced
in the subsections above rather than the ``win'' itself.
In this case, omitting the contribution of the root to the objective value
and its gradient makes the most sense, ensuring that a flat hierarchy
(that is, a hierarchy which has only one level aside from the root's)
results in the same training as with the usual cross-entropy loss.
Taking the logarithm also makes sense because the joint probability
of stochastically independent samples is the product of the probabilities
of the individual samples, making averaging across the different samples
the logarithm of a function (the function could be the win) make more sense
than averaging the function directly. That said,
taking the logarithm emphasizes highly misclassified samples,
which may not be desirable if misclassifying a few samples
(while simultaneously reporting high confidence in their classification)
should be acceptable.

Indeed, if the logarithm of the win for even a single sample is infinite,
then the average of the logarithm of the win is also infinite,
irrespective of the values for other samples.
Whether the hierarchy is full or flat, training on the logarithms of wins
is very stringent, whereas the wins without the logarithms can be
more meaningful as metrics of success or figures of merit.
It can make good sense to train on the logarithm, which works really hard
to accommodate and learn from the samples which are hardest to classify
correctly, but to make the metric of success or figure of merit be robust
against such uninteresting outliers.
Thus, training with the logarithm of the win can make good sense,
where the win --- without the logarithm --- is the metric of success
or figure of merit for the testing or validation stage.

With or without a logarithm,
we henceforth omit the contribution of the root to the hierarchical wins
and losses that we report, and we multiply by 2 the resulting win,
so that its minimal and maximal possible values become 0 and 1
(with 0 corresponding to the most incorrect possible classification,
and with 1 corresponding to the completely correct classification).

\section*{Results and discussion}
\label{numex}

We illustrate the hierarchical loss and its performance
using supervised learning for text classification with fastText
of~\cite{joulin-grave-bojanowski-mikolov}.
The tables report performance across several experiments,
with the columns ``training loss (rate, epochs)''
listing the following three parameters for training:
(1) the form of the loss function used during training (as explained shortly),
(2) the initial learning rate which tapers linearly to 0 over the course
of training, and (3) the total number of sweeps through the data performed
during training (too many sweeps results in overfitting).
The training loss ``flat'' refers to training using
the existing standard cross-entropy loss
(that is invariably the default in machine learning),
which is the same as the negative
of the natural logarithm of the hierarchical win when using a flat hierarchy
in which all labelable classes are leaves attached to the root
(as discussed in Subsection~\Ref{softmax} of Section~\Ref{methods}).
The training loss ``raw'' refers to training
using the hierarchical loss, using the full hierarchy.
The training loss ``log'' refers to training
using the negative of the natural logarithm of the hierarchical win,
using the full hierarchy.
The training loss ``coarse'' refers to training
using the usual cross-entropy loss for classification
into only the coarsest (aggregated) classes in the hierarchy
(based on a suitably smaller softmax for input to the loss).
The values reported in the tables for the learning rate and number of epochs
yielded among the best results for the accuracies discussed
in the following paragraphs, during a limited grid search
around the reported values (if increasing the epochs beyond the reported value
had little effect, then we report the smaller number of epochs);
the appendix details the settings considered.

The columns ``one-hot win via hierarchy'' display the average
over all testing samples of the hierarchical win fed with the results
of a one-hot encoding of the class chosen according to Subsection~\Ref{inbest}
of Section~\Ref{methods}.
(The one-hot encoding of a class is the vector whose entries are all zeros
aside from a single entry of one in the position corresponding to the class.)
The columns ``softmax win via hierarchy'' display the average
over all testing samples of the hierarchical win fed with the results
of a softmax from fastText of~\cite{joulin-grave-bojanowski-mikolov}
(Subsection~\Ref{softmax} of Section~\Ref{methods} above reviews
the definition of ``softmax'').
The columns ``$-$log of win via hierarchy'' display the average
over all testing samples of the negative of the natural logarithm
of the hierarchical win fed with the results of a softmax from fastText
(Subsection~\Ref{softmax} of Section~\Ref{methods} above reviews
the definition of ``softmax'').
The columns ``cross-entropy'' display the average over all testing samples
of the usual cross-entropy loss,
which is the same as the negative of the natural logarithm
of the hierarchical win fed with the results of a softmax
when using a ``flat'' hierarchy in which all labelable classes are leaves
attached to the root (as discussed in Subsection~\Ref{softmax}
of Section~\Ref{methods}).
Please note that the hierarchy and form of the training loss may be different
than the hierarchies and forms of the testing losses and wins;
for example, using the ``flat'' training loss does not alter the full hierarchy
used in the ``$-$log of win via hierarchy.''

The columns ``coarsest accuracy'' display the fraction of testing samples
for which the coarsest classes containing the fine classes chosen
during the classification are correct,
when classifying each sample into exactly one class,
as in Subsection~\Ref{inbest} of Section~\Ref{methods}.
The columns ``parents' accuracy''
display the fraction of testing samples for which the parents
of the classes chosen during the classification are correct,
when classifying each sample into exactly one class;
the parents are the same as the coarsest classes in Tables~\ref{yahoo}
and~\ref{dbpedia}, as the experiments reported in Tables~\ref{yahoo}
and~\ref{dbpedia} pertain to hierarchies with only two levels
(excluding the root).
The columns ``finest accuracy'' display the fraction of testing samples
classified correctly, when classifying each
into exactly one finest-level class,
again as in Subsection~\Ref{inbest} of Section~\Ref{methods}.

The columns ``aggregate precision,'' ``aggregate recall,''
and ``aggregate F1'' refer to the hierarchical precision, hierarchical recall,
and hierarchical F score detailed in the survey of~\cite{silla-freitas}.
The columns ``aggregate precision'' display the total number of nodes
on the path from the root to the class chosen when classifying each sample
into exactly one class, as in Subsection~\Ref{inbest} of Section~\Ref{methods},
that are also on the path from the root to the correct class,
divided by the total number of nodes on the path from the root
to the chosen class (aggregating both totals separately over all samples).
The columns ``aggregate recall'' display the same total as
for the aggregate precision, but divided by the total number of nodes
on the path from the root to the correct class, rather than on the path
from the root to the chosen class (while again aggregating both totals
in the quotient separately over all samples).
The columns ``aggregate F1'' display the harmonic mean
of the aggregate precision and the aggregate recall.
When counting these total numbers of nodes, we never count the root node,
but always count any leaf node on the paths. Since the aggregation
in the quotients defining these metrics happens separately
for each numerator and denominator, these aggregate metrics are not mean values
averaged over the samples, unlike all other quantities reported in the tables.
As elaborated by~\cite{silla-freitas}, these aggregate metrics are common
in the study of hierarchical classification (however,
hierarchical classification is beyond the scope of the present paper,
for the reasons outlined in Section~\Ref{intro};
the present paper reports the aggregate metrics at the request of some readers,
purely for reference and comparative purposes).

The last lines of the tables remind the reader that the best classifier would
have
\begin{itemize}
\item higher one-hot win via hierarchy
\item higher softmax win via hierarchy
\item lower $-$log of win via hierarchy
\item lower cross-entropy
\item higher coarsest accuracy
\item higher parents' accuracy
\item higher finest accuracy
\item higher aggregate precision
\item higher aggregate recall
\item higher aggregate F1
\end{itemize}

We follow the recommendation from Subsection~\Ref{softmax}
of Section~\Ref{methods} above, maximizing the hierarchical win
(or its logarithm) calculated without any contribution from the root
(excluding the root makes a small difference when taking the logarithm);
please note the very last paragraph in the preceeding section
--- the hierarchical wins reported here omit the contribution from the root
and rescale the result by a factor of 2.

We hashed bigrams (pairs of words) into a million buckets
and trained using stochastic gradient descent, setting the learning rate
to start at the values indicated in the tables
(these worked at least as well as other settings),
then decaying linearly to 0 over the course of training,
as done by~\cite{joulin-grave-bojanowski-mikolov};
the starting point for the stochastic gradient descent was random.
The ``learning rate'' is also known as the ``step size'' or ``step length''
in the update for each iteration (step) of stochastic gradient descent.

Our implementation couples the C++ software
of~\cite{joulin-grave-bojanowski-mikolov} with a Python prototype.
Industrial deployment requires acceleration of the Python prototype
(rewriting in C++, for instance), but our tests are sufficient
for estimating the ensuing gains in accuracy
and illustrating the figure of merit, providing a proof of principle.
In particular, our experiments indicate that the gains in accuracy
due to training with the hierarchical loss are meager
except in special circumstances detailed in the subsections below
and summarized in the conclusion.
Pending further development as suggested in the conclusion,
the main present use for the hierarchical win should be as a metric of success
or figure of merit --- a good notion of ``accuracy'' --- at least when training
with plain stochastic gradient descent coupled to backpropagation.

Several of our datasets consider the setting in which each class
includes at most one training sample. Such a setting is a proxy
for applications in ``open-world classification,'' ``few-shot learning,''
and ``personalization,'' which often involves limited
or even (for many classes) no training data per individual class,
though there may be many individuals. Access to a hierarchy of the classes
enables meaningful classification even for classes not represented
in the training data, as the hierarchy itself indicates that some classes
are very much like others (some of which may appear in the training data).
Applying a hierarchical loss is perhaps the simplest method
for effective classification and quantification of accuracy
when there is at most one training sample per class.

When reading the following subsections, please recall
that the standard cross-entropy loss is the same as the negative
of the logarithm of the hierarchical win with a flat hierarchy.

The following subsections detail our experiments and datasets;
the subsequent section, Section~\Ref{conclusion},
elucidates the consequences of these experiments.
For the experiments, we used the hierarchies that accompanied the datasets;
all are available at the web links (URLs) given in the following subsections.
Two datasets --- those in Subsections~\Ref{yahoosec} and~\Ref{dbpediasec} ---
provided no hierarchy, so for these we constructed the hierarchies
described below in Subsections~\Ref{yahoosec} and~\Ref{dbpediasec}.
In all cases, the results display little dependence on the dimension
of the space into which we embedded words and their bigrams
(words are unigrams),
in accordance with~\cite{joulin-grave-bojanowski-mikolov},
provided only that the dimension is somewhat larger than the number of classes
for targets in the classification.

\subsection*{RCV1-v2}
\label{rcv1-v2sec}

Table~\ref{rcv1-v2} reports results on RCV1-v2
of~\cite{lewis-yang-rose-li}, which is available for download
at \url{http://jmlr.csail.mit.edu/papers/volume5/lewis04a}.
This dataset includes a hierarchy
of 364 classes (semantically, these are classes of industries);
each sample from the dataset comes associated with at least one
of these 364 class labels, whether or not the class is an internal node
of the tree or a leaf.
Each sample from the dataset consists of filtered, tokenized text
from Reuters news articles (``article'' means the title and body text).
As described by~\cite{lewis-yang-rose-li},
labels associated with internal nodes in the original hierarchy may be viewed
as leaves that fall under those internal nodes while not classifying
into any of the lower-level nodes.
In our hierarchy, we hence duplicate every internal node
into an ``other'' class under that node, such that the ``other'' class
is a leaf.

We discard every sample from the dataset associated with more than one label,
and swap the training and testing sets (since the original training set
is small, whereas the original testing set is large).
Furthermore, we randomly permute all samples in both the training and testing
sets, and subsample to 5,000 samples for testing and 200,000 for training.
In the hierarchy, there are 10 coarsest classes
and 61 parents of the 254 leaves that were actually represented by any samples
in the training and testing sets.
The space into which we embedded the words and their pairs (bigrams) was
1,000-dimensional.
For training the classifier into only the coarsest (aggregated) classes,
we embedded the words and bigrams into a 20-dimensional space.

For this dataset, optimizing based on the hierarchical loss
(with or without a logarithm) yields worse accuracy according to all metrics
considered compared to optimizing based on the standard cross-entropy loss.

\begin{table}
\caption{Results on RCV1-v2, tested on 5,000 samples}
\label{rcv1-v2}
\begin{center}
\begin{tabular}{l|l|l|l|l}
\thickhline
training loss  &   one-hot win &   softmax win & $-$log of win &   cross \\
(rate, epochs) & via hierarchy & via hierarchy & via hierarchy & entropy \\
\thickhline
flat (2, 4) & .85 & .80 & .52 & .95 \\
\hline
raw (12, 4) & .51 & .50 & 4.2 & \OF \\
\hline
 log (4, 4) & .74 & .72 & .76 & 4.4 \\
\thickhline
 the ideal & higher & higher & lower & lower \\
\thickhline
\end{tabular}

\smallskip

\begin{tabular}{l|l|l|l}
\thickhline
training loss  & coarsest & parents' &   finest \\
(rate, epochs) & accuracy & accuracy & accuracy \\
\thickhline
      flat (2, 4) & .88 & .82 & .80 \\
\hline
      raw (12, 4) & .74 & .26 & .21 \\
\hline
       log (4, 4) & .87 & .63 & .53 \\
\hline
coarse (.05, 100) & .88 &     &     \\
\thickhline
 the ideal & higher & higher & higher \\
\thickhline
\end{tabular}

\smallskip

\begin{tabular}{l|l|l|l}
\thickhline
training loss  & aggregate & aggregate & aggregate \\
(rate, epochs) & precision &    recall &        F1 \\
\thickhline
      flat (2, 4) & .84 & .84 & .84 \\
\hline
      raw (12, 4) & .38 & .46 & .42 \\
\hline
       log (4, 4) & .67 & .70 & .69 \\
\thickhline
 the ideal & higher & higher & higher \\
\thickhline
\end{tabular}
\end{center}
\end{table}

\begin{table}
\caption{Results on RCV1-v2 with at most one training sample per class,
tested on 5,000 samples}
\label{rcv1s}
\begin{center}
\begin{tabular}{l|l|l|l|l}
\thickhline
training loss  &   one-hot win &   softmax win & $-$log of win &   cross \\
(rate, epochs) & via hierarchy & via hierarchy & via hierarchy & entropy \\
\thickhline
flat (.04, 9) & .14 & .07 & 2.8 & 5.5 \\
\hline
raw (45, 500) & .16 & .15 & 7.2 &  13 \\
\hline
  log (3, 40) & .17 & .09 & 2.7 & 5.5 \\
\thickhline
   the ideal & higher & higher & lower & lower \\
\thickhline
\end{tabular}

\smallskip

\begin{tabular}{l|l|l|l}
\thickhline
training loss  & coarsest & parents' &   finest \\
(rate, epochs) & accuracy & accuracy & accuracy \\
\thickhline
   flat (.04, 9) & .209 & .086 & .051 \\
\hline
   raw (45, 500) & .235 & .095 & .052 \\
\hline
     log (3, 40) & .258 & .089 & .064 \\
\hline
coarse (4, 1000) & .324 &      &      \\
\thickhline
   the ideal & higher & higher & higher \\
\thickhline
\end{tabular}

\smallskip

\begin{tabular}{l|l|l|l}
\thickhline
training loss  & aggregate & aggregate & aggregate \\
(rate, epochs) & precision &    recall &        F1 \\
\thickhline
   flat (.04, 9) & .121 & .118 & .120 \\
\hline
   raw (45, 500) & .132 & .148 & .140 \\
\hline
     log (3, 40) & .141 & .166 & .153 \\
\thickhline
 the ideal & higher & higher & higher \\
\thickhline
\end{tabular}
\end{center}
\end{table}

\subsection*{Subsampled RCV1-v2}
\label{rcv1ssec}

Table~\ref{rcv1s} reports results on the same dataset RCV1-v2
(which is available for download
at \url{http://jmlr.csail.mit.edu/papers/volume5/lewis04a})
of the preceeding subsection, but retaining only one training sample
for each class label. The training set thus consists of 254 samples
(many of the 364 possible class labels had no corresponding samples
in the training set from the preceeding subsection).
As in the preceding subsection, the hierarchy has 10 coarsest classes
and 61 parents of the 254 leaves that were used
in the training and testing sets.
We again embedded the words and their bigrams into a 1,000-dimensional space,
while for training the classifier into only the coarsest (aggregated) classes,
we again used only 20 dimensions.

For this subsampled RCV1-v2,
optimizing based on the negative of the natural logarithm
of the hierarchical win yields better accuracy according to all metrics
considered compared to optimizing based on the standard cross-entropy loss,
except on the negative of the natural logarithm of the hierarchical win
and the cross-entropy loss (for which the accuracies are similar).

\begin{table}
\caption{Results on Yahoo Answers, tested on 60,000 samples}
\label{yahoo}
\begin{center}
\begin{tabular}{l|l|l|l|l}
\thickhline
training loss  &   one-hot win &   softmax win & $-$log of win &   cross \\
(rate, epochs) & via hierarchy & via hierarchy & via hierarchy & entropy \\
\thickhline
flat (.1, 4) & .76 & .67 & .67 & .91 \\
\hline
 raw  (1, 4) & .65 & .64 & 2.6 & \OF \\
\hline
 log (.1, 4) & .76 & .67 & .68 & 1.0 \\
\thickhline
  the ideal & higher & higher & lower & lower \\
\thickhline
\end{tabular}

\smallskip

\begin{tabular}{l|l|l}
\thickhline
training loss  & coarsest &   finest \\
(rate, epochs) & accuracy & accuracy \\
\thickhline
  flat (.1, 4) & .80 & .72 \\
\hline
   raw  (1, 4) & .79 & .50 \\
\hline
   log (.1, 4) & .80 & .71 \\
\hline
coarse (.1, 4) & .80 &     \\
\thickhline
  the ideal & higher & higher \\
\thickhline
\end{tabular}

\smallskip

\begin{tabular}{l|l|l|l}
\thickhline
training loss  & aggregate & aggregate & aggregate \\
(rate, epochs) & precision &    recall &        F1 \\
\thickhline
  flat (.1, 4) & .76 & .76 & .76 \\
\hline
   raw  (1, 4) & .64 & .64 & .64 \\
\hline
   log (.1, 4) & .76 & .76 & .76 \\
\thickhline
 the ideal & higher & higher & higher \\
\thickhline
\end{tabular}
\end{center}
\end{table}

\subsection*{Yahoo Answers}
\label{yahoosec}

Table~\ref{yahoo} reports results on the Yahoo Answers subset
introduced by~\cite{zhang-zhao-lecun}, which is available for download
at \url{http://goo.gl/JyCnZq}. This dataset includes 10 classes
(semantically, these are classes of interest groups);
each sample from the dataset comes associated with exactly one
of these 10 class labels.
Each sample from the dataset consists of normalized text from questions
and answers given on a website devoted to Q\&A.
The space into which we embedded the words and their bigrams was
20-dimensional, while for training the classifier
into only the coarsest (aggregated) classes,
we embedded the words and bigrams into a 10-dimensional space.
For the nontrivial hierarchy, we grouped the 10 classes into 4 superclasses:
\begin{description}
\item[1) Leisure:] Entertainment and Music, Society and Culture, and Sports
\item[2) Newsworthy:] Business and Finance, and Politics and Government
\item[3) Relations:] Family and Relationships
\item[4) Science and Technology:] Computer and Internet,
                                  Education and Reference, Health,
                                  and Science and Mathematics
\end{description}

With only 10 classes and two levels for the classification hierarchy,
Table~\ref{yahoo} indicates that training with or without the hierarchical loss
makes little difference.

\begin{table}
\caption{Results on DBpedia, tested on 70,000 samples}
\label{dbpedia}
\begin{center}
\begin{tabular}{l|l|l|l|l}
\thickhline
training loss  &   one-hot win &   softmax win & $-$log of win &   cross \\
(rate, epochs) & via hierarchy & via hierarchy & via hierarchy & entropy \\
\thickhline
flat (.5, 4) & .989 & .986 & .034 & .054 \\
\hline
 raw  (1, 4) & .813 & .811 & .312 & 6.06 \\
\hline
 log (.5, 4) & .988 & .985 & .036 & .063 \\
\thickhline
  the ideal & higher & higher & lower & lower \\
\thickhline
\end{tabular}

\smallskip

\begin{tabular}{l|l|l}
\thickhline
training loss  & coarsest &   finest \\
(rate, epochs) & accuracy & accuracy \\
\thickhline
  flat (.5, 4) & .992 & .986 \\
\hline
   raw  (1, 4) & .989 & .636 \\
\hline
   log (.5, 4) & .992 & .985 \\
\hline
coarse (.3, 4) & .992 &      \\
\thickhline
  the ideal & higher & higher \\
\thickhline
\end{tabular}

\smallskip

\begin{tabular}{l|l|l|l}
\thickhline
training loss  & aggregate & aggregate & aggregate \\
(rate, epochs) & precision &    recall &        F1 \\
\thickhline
  flat (.5, 4) & .989 & .989 & .989 \\
\hline
   raw  (1, 4) & .813 & .813 & .813 \\
\hline
   log (.5, 4) & .988 & .988 & .988 \\
\thickhline
 the ideal & higher & higher & higher \\
\thickhline
\end{tabular}
\end{center}
\end{table}

\subsection*{DBpedia}
\label{dbpediasec}

Table~\ref{dbpedia} reports results on the DBpedia subset
introduced by~\cite{zhang-zhao-lecun}, which is available for download
at \url{http://goo.gl/JyCnZq}. This dataset includes 14 classes
(semantically, these are categories from DBpedia);
each sample from the dataset comes associated with exactly one
of these 14 class labels.
Each sample from the dataset consists of normalized text from DBpedia articles
(``article'' means the title and body text). 
We embedded the words and their bigrams into a 20-dimensional space,
except when training a classifier into only the coarsest (aggregated) classes,
in which case a space with only 10 dimensions was fine.
For the nontrivial hierarchy, we grouped the 14 classes into 6 superclasses:
\begin{description}
\item[1) Institution:] Company and EducationalInstitution
\item[2) Man-made:] Building and MeansOfTransportation
\item[3) Media:] Album, Film, and WrittenWork
\item[4) Organism:] Animal and Plant
\item[5) Person:] Artist, Athlete, and OfficeHolder
\item[6) Place:] NaturalPlace and Village
\end{description}

With merely 14 classes and two levels for the classification hierarchy,
Table~\ref{dbpedia} shows that training
with or without the hierarchical loss makes little difference.

\begin{table}
\caption{Results on DBpedia fish, tested on 6,000 samples}
\label{dbpediafish}
\begin{center}
\begin{tabular}{l|l|l|l|l}
\thickhline
training loss  &   one-hot win &   softmax win & $-$log of win &   cross \\
(rate, epochs) & via hierarchy & via hierarchy & via hierarchy & entropy \\
\thickhline
  flat (5, 25) & .27 & .25 & 4.4 &  16 \\
\hline
raw (200, 200) & .17 & .17 & 6.5 & \OF \\
\hline
   log (5, 50) & .38 & .36 & 3.5 &  19 \\
\thickhline
   the ideal & higher & higher & lower & lower \\
\thickhline
\end{tabular}

\smallskip

\begin{tabular}{l|l|l|l}
\thickhline
training loss  & coarsest & parents' &   finest \\
(rate, epochs) & accuracy & accuracy & accuracy \\
\thickhline
  flat (5, 25) & .38 & .26 & .065 \\
\hline
raw (200, 200) & .31 & .04 & .001 \\
\hline
   log (5, 50) & .60 & .22 & .014 \\
\hline
coarse (3, 70) & .60 &     &      \\
\thickhline
   the ideal & higher & higher & higher \\
\thickhline
\end{tabular}

\smallskip

\begin{tabular}{l|l|l|l}
\thickhline
training loss  & aggregate & aggregate & aggregate \\
(rate, epochs) & precision &    recall &        F1 \\
\thickhline
  flat (5, 25) & .24 & .25 & .24 \\
\hline
raw (200, 200) & .12 & .13 & .12 \\
\hline
   log (5, 50) & .30 & .33 & .31 \\
\thickhline
 the ideal & higher & higher & higher \\
\thickhline
\end{tabular}
\end{center}
\end{table}

\subsection*{DBpedia fish}
\label{fishsec}

Table~\ref{dbpediafish} reports results on the subset corresponding to fish
from the DBpedia of~\cite{lehmann-et_al}, which is available for download at
\url{http://web.informatik.uni-mannheim.de/DBpediaAsTables/DBpedia-3.9/json/Fish.json.gz}.
This dataset includes 1,298 classes
(semantically, these are taxonomic groups of fish, such as species containing
sub-species, genera containing species, or families containing genera
--- DBpedia extends to different depths of taxonomic rank for different kinds
of fish; our classes are the parents of the leaves in the DBpedia tree).
Each sample from the dataset consists of normalized text from the lead section
(the introduction) of the Wikipedia article on the associated type of fish,
with all sub-species, species, genus, family, and order names
removed from the associated Wikipedia article (DBpedia derives from Wikipedia,
as discussed by~\cite{lehmann-et_al}).
For each of our finest-level classes, we chose uniformly at random one leaf
in the DBpedia taxonomic tree of fish to be a sample in the training set,
reserving the other leaves for the testing set (the testing set consists
of a random selection of 6,000 of these leaves).
In the hierarchy, there are 94 coarsest classes,
367 parents of the leaves in our tree, and 1,298 leaves in our tree.
Optimizing the hierarchical win --- without any logarithm ---
was wholly ineffective,
always resulting in assigning the same finest-level class to all input samples
(with the particular class assigned varying according to the extent
of training and the random starting point).
So taking the logarithm of the hierarchical win was absolutely necessary
to train successfully.
For training the classifier into only the coarsest (aggregated) classes,
we embedded the Wikipedia articles' words and bigrams
into a 200-dimensional space rather than the larger 2,000-dimensional space
used for classifying into all 1,298 classes.

Here, optimizing based on the negative of the logarithm
of the hierarchical win yields much better coarsest accuracy
and hierarchical wins than optimizing based on the standard cross-entropy loss,
while optimizing based on the standard cross-entropy loss
yields much better finest accuracy and cross-entropy.
When optimizing based on the negative of the natural log
of the hierarchical win, the accuracy on the coarsest aggregates
reaches that attained when optimizing the coarse classification directly.

\begin{table}
\caption{Results on a subset of LSHTC1, tested on 2,000 samples}
\label{lshtc1}
\begin{center}
\begin{tabular}{l|l|l|l|l}
\thickhline
training loss  &   one-hot win &   softmax win & $-$log of win &   cross \\
(rate, epochs) & via hierarchy & via hierarchy & via hierarchy & entropy \\
\thickhline
   flat (6, 1000) & .43 & .36 & 1.6 & 5.0 \\
\hline
raw (20000, 1000) & .23 & .23 & .35 & \OF \\
\hline
     log (15, 15) & .36 & .30 & 1.7 & 5.5 \\
\thickhline
   the ideal & higher & higher & lower & lower \\
\thickhline
\end{tabular}

\smallskip

\begin{tabular}{l|l|l|l}
\thickhline
training loss  & coarsest & parents' &   finest \\
(rate, epochs) & accuracy & accuracy & accuracy \\
\thickhline
   flat (6, 1000) & .61 & .33 & .23 \\
\hline
raw (20000, 1000) & .46 & .24 & .01 \\
\hline
     log (15, 15) & .66 & .12 & .01 \\
\hline
coarse (5, 10000) & .69 &     &     \\
\thickhline
   the ideal & higher & higher & higher \\
\thickhline
\end{tabular}

\smallskip

\begin{tabular}{l|l|l|l}
\thickhline
training loss  & aggregate & aggregate & aggregate \\
(rate, epochs) & precision &    recall &        F1 \\
\thickhline
   flat (6, 1000) & .38 & .39 & .38 \\
\hline
raw (20000, 1000) & .23 & .18 & .20 \\
\hline
     log (15, 15) & .26 & .29 & .28 \\
\thickhline
 the ideal & higher & higher & higher \\
\thickhline
\end{tabular}
\end{center}
\end{table}

\subsection*{LSHTC1}
\label{lshtcsec}

Table~\ref{lshtc1} reports results on a subset of the LSHTC1 dataset
introduced by~\cite{lshtc}, which is available for download
at \url{http://lshtc.iit.demokritos.gr} (specifically, ``DMOZ large''
for ``LSHTC1''). The subset considered consists of the subtree
for class 3261; this subtree includes 18 coarsest classes (though 3
of these have no corresponding samples in the testing or training sets)
and 364 finest-level classes (with 288 of these having corresponding samples in
the testing and training sets). We reserved one sample per finest-level class
for training; all other samples were for testing, and we chose 2,000 of these
uniformly at random to form the testing set. Each sample from the dataset
consists of normalized, tokenized text in extracts from Wikipedia,
the popular crowdsourced online encyclopedia.
The hierarchy has 18 coarsest classes
(with 15 actually represented in the training and testing sets),
as well as 111 parents of the 288 leaves that were used
in the training and testing sets.
We embedded the words and their bigrams into a 1,000-dimensional space.
For training the classifier into only the coarsest classes,
we embedded the words and bigrams into a 20-dimensional space.

For this dataset, optimizing based on the hierarchical loss
(with or without a logarithm) yields worse accuracy according to all metrics
except the accuracy on the coarsest aggregates,
compared to optimizing based on the standard cross-entropy loss.
When optimizing based on the negative of the natural logarithm
of the hierarchical win, the accuracy on the coarsest aggregates approaches
its maximum attained when optimizing the coarse classification directly.

\section*{Conclusion}
\label{conclusion}

In our experiments, optimizing the hierarchical loss
(or, rather, the negative of the logarithm of the hierarchical win) using plain
stochastic gradient descent with backpropagation could be helpful
relative to optimizing the usual cross-entropy loss.
The benefit arose mainly when there were at most a few training samples
per class (of course, the training set can still be big,
if there are many classes).
This may help with ``personalization,'' which often involves limited data
per individual class (even though there may be many, many individuals).

The experiments reported in Section~\Ref{numex} above may be summarized
as follows: relative to training on the usual cross-entropy loss,
training on the negative of the logarithm of the hierarchical win
hurt in all respects in Table~\ref{rcv1-v2},
helped in all respects in Table~\ref{rcv1s},
improved coarse accuracy as much as optimizing directly
for coarse classification in Table~\ref{dbpediafish},
hurt in most respects in Table~\ref{lshtc1} while improving coarse accuracy
nearly as much as optimizing directly for coarse classification,
and made essentially no difference in Tables~\ref{yahoo}
and~\ref{dbpedia}. Thus, whether optimizing with a hierarchical loss
makes sense depends on the dataset and associated hierarchy;
developing precise criteria for when the optimization
with the hierarchical loss helps might be interesting.

Even so, optimizing hierarchical loss using plain stochastic gradient descent
with backpropagation (as we did) is rather ineffective, at least relative
to what might be possible. We trained using stochastic gradient descent
with a random starting point, which may be prone to getting stuck
in local optima. To some extent, hierarchical loss collapses the many classes
in the hierarchy into a few aggregate superclasses, and the parameters
being optimized within the aggregates should be tied closely together
during the optimization --- plain stochastic gradient descent is unlikely
to discover the benefits of such tying, as plain stochastic gradient descent
does not tie together these parameters in any way, optimizing all of them
independently. Optimizing the hierarchical loss would presumably be more
effective using a hierarchical process for the optimization.
The hierarchical optimization could alter stochastic gradient descent
explicitly into a hierarchical process, or could involve regularization terms
penalizing variance in the parameters associated with the leaves
in the same coarse aggregate. Unfortunately, either approach would complicate
(perhaps prohibitively much) deployment to complex real-world systems.
For the time being, hierarchical loss is most useful as a metric of success,
gauging the performance of a fully trained classifier
as a semantically meaningful figure of merit.
While the performance of the hierarchical loss as an objective for optimization
varies across datasets, the hierarchical loss provides a meaningful metric
of success for classification on any dataset endowed with a hierarchy.

\section*{Acknowledgements}

We would like to thank Priya Goyal, Anitha Kannan, and Brian Karrer
for their generous and plentiful help.

\appendix
\section*{Appendix: Minutiae of tuning hyperparameters}

Table~\ref{grids} elaborates on the aforementioned limited grid search
performed to ascertain reasonably optimal learning rates and numbers of epochs
for training.

\begin{table}
\caption{Hyperparameters for grid search}
\label{grids}
\begin{center}
\begin{tabular}{l|l|l|l}
\thickhline
dataset & loss & learning rates & epochs \\
\thickhline
RCV1-v2 & flat & 1, 2, 4 & 3, 4, 5, 6 \\
RCV1-v2 & raw & 10, 12, 16 & 3, 4, 5, 6 \\
RCV1-v2 & log & 2, 4, 6 & 3, 4, 5, 6 \\
RCV1-v2 & coarse & .03, .05, .07 & 50, 100, 200 \\
\thickhline
Subsampled RCV1-v2 & flat & .02, .03, .04, .05 & 8, 9, 10, 11 \\
Subsampled RCV1-v2 & raw & 40, 45, 50 & 300, 500, 600 \\
Subsampled RCV1-v2 & log & 1, 2, 3, 4 & 30, 40, 50 \\
Subsampled RCV1-v2 & coarse & 2, 4, 8 & 500, 1000, 2000 \\
\thickhline
Yahoo Answers & flat & .05, .1, .2, .4 & 2, 4, 8, 16 \\
Yahoo Answers & raw & .5, 1, 1.5 & 2, 4, 8, 16 \\
Yahoo Answers & log & .05, .1, .2 & 2, 4, 8, 16 \\
Yahoo Answers & coarse & .05, .1, .2 & 2, 4, 8 \\
\thickhline
DBpedia & flat & .125, .25, .5, 1, 2 & 2, 4, 8 \\
DBpedia & raw & .5, 1, 2, 4 & 2, 4, 8 \\
DBpedia & log & .125, .25, .5, 1, 2 & 2, 4, 8 \\
DBpedia & coarse & .1, .2, .3, .4 & 2, 4, 8 \\
\thickhline
DBpedia fish & flat & 3, 5, 10 & 20, 25, 30 \\
DBpedia fish & raw & 100, 200, 300 & 100, 200, 300 \\
DBpedia fish & log & 3, 5, 10 & 25, 50, 100 \\
DBpedia fish & coarse & 1, 2, 3, 4 & 50, 60, 70, 80 \\
\thickhline
LSHTC1 & flat & 2, 4, 6, 8 & 500, 1000, 2000 \\
LSHTC1 & raw & 10000, 20000, 30000 & 500, 1000, 2000 \\
LSHTC1 & log & 10, 15, 20 & 10, 15, 20 \\
LSHTC1 & coarse & 3, 5, 7 & 5000, 10000, 20000 \\
\thickhline
\end{tabular}
\end{center}
\end{table}

\clearpage



\begin{thebibliography}{10}

\bibitem{lecun-bengio-hinton}
Le{C}un Y, Bengio Y, Hinton G.
\newblock Deep learning.
\newblock Nature. 2015;521(7553):436--444.

\bibitem{cai-hofmann}
Cai L, Hofmann T.
\newblock Hierarchical document categorization with support vector machines.
\newblock In: Proc.\ 13th ACM Internat.\ Conf.\ Information and Knowledge
  Management. ACM; 2004. p. 78--87.

\bibitem{kosmopoulos-partalas-gaussier-paliouras-androutsopoulos}
Kosmopoulos A, Partalas I, Gaussier E, Paliouras G, Androutsopoulos I.
\newblock Evaluation measures for hierarchical classification: a unified view
  and novel approaches.
\newblock Data Mining and Knowledge Discovery. 2015;29(3):820--865.

\bibitem{binder-kawanabe-brefeld}
Binder A, Kawanabe M, Brefeld U.
\newblock Efficient classification of images with taxonomies.
\newblock In: Proc.\ 9th Asian Conf.\ Computer Vision. vol. 5996 of Lecture
  Notes in Computer Science. Springer; 2009. p. 351--362.

\bibitem{chang-lee}
Chang JY, Lee KM.
\newblock Large margin learning of hierarchical semantic similarity for image
  classification.
\newblock Computer Vision and Image Understanding. 2015;132:3--11.

\bibitem{costa-lorena-carvalho-freitas}
Costa EP, Lorena AC, Carvalho ACPLF, Freitas AA.
\newblock A review of performance evaluation measures for hierarchical
  classifiers.
\newblock In: Drummond C, Elazmeh W, Japkowicz N, Macskassy SA, editors.
  Evaluation Methods for Machine Learning II: Papers from the AAAI-2007
  Workshop. AAAI Press; 2007. p. 182--196.

\bibitem{deng-berg-li-li1}
Deng J, Berg AC, Li K, Li FF.
\newblock What does classifying more than 10,000 image categories tell us?
\newblock In: Proc.\ 11th European Conf.\ Computer Vision. vol.~5.
  Springer-Verlag; 2010. p. 71--84.

\bibitem{deng-berg-li-li2}
Deng J, Berg AC, Li K, Li FF.
\newblock Hierarchical semantic indexing for large scale image retrieval.
\newblock In: Proc.\ IEEE Conf.\ Computer Vision and Pattern Recognition. IEEE;
  2011. p. 785--792.

\bibitem{wang-zhou-liew}
Wang K, Zhou S, Liew SC.
\newblock Building hierarchical classifiers using class proximity.
\newblock In: Proc.\ 25th Internat.\ Conf.\ Very Large Data Bases. Morgan
  Kaufmann Publishers; 1999. p. 363--374.

\bibitem{campbell}
Reece JB, Urry LA, Cain ML, Wasserman SA, Minorsky PV, Jackson RB.
\newblock Campbell Biology.
\newblock 10th ed. Pearson; 2013.

\bibitem{silla-freitas}
{Silla, Jr } CN, Freitas AA.
\newblock A survey of hierarchical classification across different application
  domains.
\newblock J.\ Data Mining Knowledge Discovery. 2011;22(1--2):31--72.

\bibitem{kosmopoulos-paliouras-androutsopoulos}
Kosmopoulos A, Paliouras G, Androutsopoulos I.
\newblock Probabilistic cascading for large-scale hierarchical classification.
\newblock arXiv; 2015. 1505.02251.
\newblock Available from: \url{http://arxiv.org/abs/1505.02251}.

\bibitem{redmon-farhadi}
Redmon J, Farhadi A.
\newblock {YOLO9000}: better, faster, stronger.
\newblock In: IEEE Conf.\ Comput.\ Vision Pattern Recognition. IEEE; 2017. p.
  1--9.

\bibitem{joulin-grave-bojanowski-mikolov}
Joulin A, Grave E, Bojanowski P, Mikolov T.
\newblock Bag of tricks for efficient text classification.
\newblock In: Proc.\ 15th Conf.\ European Chapter Assoc.\ Comput.\ Linguistics.
  ACL; 2017. p. 427--431.

\bibitem{lewis-yang-rose-li}
Lewis DD, Yang Y, Rose TG, Li F.
\newblock {RCV1}: a new benchmark collection for text categorization research.
\newblock J.\ Machine Learning Research. 2004;5:361--397.

\bibitem{zhang-zhao-lecun}
Zhang X, Zhao J, {LeC}un Y.
\newblock Character-level convolutional networks for text classification.
\newblock In: Advances in Neural Information Processing Systems. vol.~28.
  Neural Information Processing Systems Foundation; 2015. p. 1--9.

\bibitem{lehmann-et_al}
Lehmann J, Isele R, Jakob M, Jentzsch A, Kontokostas D, Mendes PN, et~al.
\newblock {DB}pedia --- a large-scale, multilingual knowledge base extracted
  from {W}ikipedia.
\newblock Semantic Web. 2015;6(2):167--195.

\bibitem{lshtc}
Partalas I, Kosmopoulos A, Baskiotis N, Artieres T, Paliouras G, Gauss\-ier E,
  et~al.
\newblock {LSHTC}: a benchmark for large-scale text classification.
\newblock arXiv; 2015. 1503.08581.
\newblock Available from: \url{http://arxiv.org/abs/1503.08581}.

\end{thebibliography}
\end{document}